\documentclass[sigconf]{acmart}

\pdfoutput=1

\AtBeginDocument{%
  }

\setcopyright{acmcopyright}
\copyrightyear{2022}
\acmYear{2022}
\acmDOI{}

\acmConference[International Joint Conference on Knowledge Graphs ]{October 27–29, 2022}{Hangzhou, China}{}
\acmBooktitle{}
\acmPrice{}
\acmISBN{}

\usepackage[linesnumbered,ruled]{algorithm2e}
\usepackage{CJKutf8}
\usepackage{pgfplots}
\pgfplotsset{width=6.6cm,compat=1.14}

\begin{document}

\title{KG-BERTScore: Incorporating Knowledge Graph into BERTScore for Reference-Free Machine Translation Evaluation}

\author{Zhanglin Wu, Min Zhang, Ming Zhu, Yinglu Li, Ting Zhu, Hao Yang*, Song Peng, Ying Qin}
\email{{wuzhanglin2, zhangmin186, zhuming47, liyinglu, zhuting20, yanghao30, pengsong2, qinying}@huawei.com}
\affiliation{%
  \institution{Huawei Technologies Co., Ltd.}
  \city{Beijing}
  \country{China}
  \postcode{100000}
}

\begin{abstract}
BERTScore is an effective and robust automatic metric for reference-based machine translation evaluation. In this paper, we incorporate multilingual knowledge graph into BERTScore and propose a metric named KG-BERTScore, which linearly combines the results of BERTScore and bilingual named entity matching for reference-free machine translation evaluation. From the experimental results on WMT19 QE as a metric without references shared tasks, our metric KG-BERTScore gets higher overall correlation with human judgements than the current state-of-the-art metrics for reference-free machine translation evaluation.\footnote{\url{https://www.statmt.org/wmt19/qe-task.html}} Moreover, the pre-trained multilingual model used by KG-BERTScore and the parameter for linear combination are also studied in this paper.

\end{abstract}


\maketitle

\begin{table*}[ht]
\begin{center}
\small
\caption{A KG-BERTScore calculation example for en-zh. $F_{BERT}$ is calculated by xlm-roberta-base, and $\alpha$ is set to 0.5.}\label{tab1}%
\scalebox{1.25}{
\begin{tabular}{l l}
\toprule
source sentence & \textcolor{red}{Respiratory} irritation was not reported in Northwest \textcolor{red}{Florida} over the past \textcolor{red}{week}.\\
corresponding entity IDs & /m/0hl\_6 \quad \quad \quad \quad \quad \quad \quad \quad \quad \quad \quad \quad \quad \quad \quad \quad \quad \textbf{/m/02xry} \quad \quad \quad \quad /m/083sl\\
\hline
machine translation & \begin{CJK}{UTF8}{gbsn}
\textcolor{red}{本周}，\textcolor{red}{佛罗里达}西北部没有\textcolor{red}{消化道}刺激的\textcolor{red}{报告}。
\end{CJK}\\
corresponding entity IDs &  /m/05qv5f \textbf{/m/02xry} \quad \quad /m/0j49l \quad \quad \quad /m/0chln1\\
\hline
$F_{BERT}$ & 0.857\\
$F_{KG}$ & $\frac{1}{3}=0.333$\\
$F_{KG-BERT}$ & $0.333 \times 0.5+0.857 \times (1-0.5)=0.595$\\
\bottomrule
\end{tabular}}
\end{center}
\end{table*}

\section{Introduction}\label{sec1}
Machine translation (MT) evaluation is an important research topic in natural language processing, and its development plays a crucial role in the progress of machine translation. Although Human judgement is an ideal MT evaluation metric, automatic MT evaluation metrics are applied in most cases due to the former's long evaluation cycle and high labor consumption. With the continuous deepening of research, automatic MT evaluation metrics have been divided into reference-based MT evaluation metrics and reference-free MT evaluation metrics\cite{bojar2014findings}.

Reference-based MT evaluation metrics are mainly separated into three categories: n-gram similarity based-metrics, editing distance based metrics and word embedding-based metrics. N-gram similarity-based metrics include BLEU\cite{papineni2002bleu}, chrF\cite{popovic2015chrf} and METEOR \cite{banerjee2005meteor}, etc. Among these metrics, BLEU measures the correspondence of n-grams between machine translation and reference, chrF uses character n-gram instead of word n-gram, and METEOR takes into account the form of words, expands thesaurus with knowledge sources such as WordNet\cite{miller1995wordnet}. Editing distance-based metrics such as TER\cite{snover2006study}, WER\cite{och2003minimum} and PER\cite{och2003minimum} evaluate the quality based on the minimum number of edits required to convert a machine translation into a reference, among which the difference lies in the definition of "error" and the type of editing action. The above two types of reference-based metrics can only provide rule-based metrics while word embedding-based metrics can also take into account the intrinsic meaning of words. Word embedding-based metrics are divided into semantic information-based metrics and end-to-end evaluation metrics. The semantic information-based metrics use pre-training model such as word2vector or BERT\cite{kenton2019bert} to perform lexical level analysis and alignment of machine translation and reference to calculate the semantic similarity, whose implementations include MEANT 2.0\cite{lo2017meant}, YiSI\cite{lo2019yisi}, BLEURT\cite{sellam2020bleurt} and BERTScore\cite{zhang2020bertscore}, etc.\footnote{\url{https://github.com/Tiiiger/bert\_score}} The end-to-end evaluation metrics usually adopt predictor-estimator\cite{kim2017predictor} architecture and use 
multilingual pre-training model such as XLM-R\cite{conneau2020unsupervised} as predictor for encoding and estimator for scoring. These metrics such as COMET\cite{stewart2020comet}, rely on human scoring data to train the model.

Reference-free MT evaluation metrics are more challenging and promising compared with reference-based ones. Early reference-free MT evaluation metrics like QuEST\cite{specia2013quest} and QuEST++\cite{specia2015multi} are heavily dependent on linguistic processing and feature engineering to train traditional machine-learning algorithms like support vector regression and randomised decision trees\cite{specia2013quest}. Such metrics are called artificial feature based reference-free MT evaluation metrics. In effect, these metrics are generally considered to be inferior to neural network based reference-free MT evaluation metrics. Neural network based reference-free MT evaluation metrics refer to using neural network for end-to-end modeling, automatically extracting features, and then evaluating the quality of machine translation. Early neural network based reference-free MT evaluation metrics include POSTECH\cite{kim2017predictor} and deepquest\cite{ive2018deepquest}, which require pre-training with large-scale parallel corpus. In the OpenKiwi\cite{kepler-etal-2019-openkiwi} framework, there is an improved metric based on the pre-trained language model, which avoids relying on large-scale parallel corpus and requires only a small amount of human scoring data to fine-tune neural network. On this basis, TransQuest\cite{ranasinghe2020transquest} selects XLM-R as the pre-training language model, and simplifies the structure of neural network, which improves the computational efficiency and evaluation accuracy.

MT evaluation metrics focus on how to improve the correlation between evaluation results with human judgements. BERTScore is a typical reference-based MT evaluation metric, which correlates better with human judgements. It can also be used as a reference-free MT evaluation metric by embedding words using pre-trained multilingual model, but it correlates badly with human judgement. Therefore, we incorporate multilingual knowledge graph\cite{chen2017multilingual} into BERTScore for reference-free MT evaluation, and propose a ref-erence-free metric KG-BERTScore. Our metric uses multilingual knowledge graph and pre-trained multilingual model instead of fine-tuning on parallel corpora and human scoring data, which can accurately evaluate machine translations.

To summarize, our work includes the following contributions:

\begin{itemize}
\item To the best of our knowledge, we are the first to focus on how to combine multilingual knowledge graph with pre-trained multilingual model for reference-free MT evaluation.
\item We propose an unsupervised metric KG-BERTScore, which incorporates multilingual knowledge graph into BERTScore for reference-free MT evaluation.
\item We show that KG-BERTScore correlates better with human judgment on the WMT19 QE as a metric without references shared task\cite{fonseca2019findings} than the current state-of-the-art reference-free MT evaluation metrics.
\end{itemize}

\section{Methods}\label{sec2}

\subsection{BERTScore}\label{subsec1}

BERTScore is a reference-based MT evaluation metric. We try to use it for reference-free MT evaluation and the specific steps to generate a system-level score can be described as follows:

First, the word embedding is generated by pre-trained multilingual model, and then the cosine similarity $x_i^T \hat{x}_j$ of each word $x_i$ in source text and each word $\hat{x}_j$ in machine translation is calculated. We use greedy matching to maximize the cosine similarity score, where each word matches the most similar word in another sentence. 

Then, we calculate $F_{BERT}$ Score for each machine translation sentence as follows:

\begin{equation}
R=\frac{1}{\lvert x\rvert}\sum\limits_{x_i \in x}\max\limits_{\hat{x}_j \in \hat{x}}x_i^T\hat{x}_j\label{eq1}
\end{equation}
\begin{equation}
P=\frac{1}{\lvert \hat{x}\rvert}\sum\limits_{\hat{x}_i \in \hat{x}}\max\limits_{x_j \in x}\hat{x}_i^T x_j\label{eq2}
\end{equation}
\begin{equation}
F_{BERT}=2\frac{P\cdot R}{P+R}\label{eq3}
\end{equation}

Finally, we average $F_{BERT}$ Score of all machine translation sentences to obtain a system-level score.

\subsection{KG-BERTScore}\label{subsec2}

We put forward a reference-free KG-BERTScore MT evaluation metric, which incorporates multilingual knowledge graph into BERTScore for reference-free MT evaluation. The evaluation process is shown in Algorithm \ref{alg1}: 

\IncMargin{1em}
\begin{algorithm}[ht]

\KwIn{all source sentences $s_k \in S$ and machine translations $t_k \in T$ of $n$ sentence pairs}
\KwOut{a system-level score $F$}

\For{each sentence pair $\{s_k,t_k\} \in \{S,T\}$}{
    \tcp{$x_i, x_j, \hat{x}_i, \hat{x}_j$ is the word embedding.}
    
    $R_k=\frac{1}{\lvert s_k\rvert}\sum\limits_{x_i \in s_k}\max\limits_{\hat{x}_j \in t_k}x_i^T\hat{x}_j$

    $P_k=\frac{1}{\lvert t_k\rvert}\sum\limits_{\hat{x}_i \in t_k}\max\limits_{x_j \in s_k}\hat{x}_i^T x_j$

    $F_{BERT_k}=2\frac{P_k\cdot R_k}{P_k+R_k}$
    
    \tcp{$entities\left({s_k}\right)$, $entities\left({t_k}\right)$ is the number of entities.}
    
    \eIf{$entities\left({s_k}\right) \neq 0$}{
     $F_{KG_k}=\frac{matches\left( entities\left({s_k}\right),entities\left({t_k}\right)\right)}{entities\left({s_k}\right)}$
    }{
     $F_{KG_k}=1$
     }   
     
    \tcp{$\alpha$ is an adjustable hyperparameter.}
    
     $F_{KG-BERT_k}=\alpha\cdot F_{KG_k}+(1-\alpha)\cdot F_{BERT_k}$
    
}

$F=\frac{\sum_{k=1}^{n}{F_{KG-BERT_k}}}{n}$

\caption{KG-BERTScore evaluation process}
\label{alg1}
\end{algorithm}
\DecMargin{1em}

Firstly, we employ reference-free BERTScore metric to calculate $F_{BERT}$ score of each machine translation sentence.

Secondly, we annotate the named entities and the corresponding entity IDs in the sentences and calculate the entity matching scores. We can utilize named entity recognition model such as W-NER\cite{yan2021unified} to identify named entities, and entity links\cite{lu2020enrich} to retrieve their entity IDs in multilingual knowledge graph. We then calculate $F_{KG}$ scores based on entity matching degree. Since the same named entities in different languages share the same entity ID in multilingual knowledge graph, we can check whether they can be matched by entity IDs. Specifically, for source sentence $s$ and machine translation sentence $t$, the $F_{KG}$ score is calculated as follows:
\begin{equation}
F_{KG}=\frac{matches\left( entities\left({s}\right),entities\left({t}\right)\right)}{entities\left({s}\right)}.\label{eq4}
\end{equation}

Then, the above two scores are combined to obtain $F_{KG-BERT}$ score as the final evaluation result of machine translation sentence.
 
 \begin{equation}
F_{KG-BERT}=\alpha\cdot F_{KG}+(1-\alpha)\cdot F_{BERT}\label{eq5}
\end{equation}

Finally, we average $F_{KG-BERT}$ score of all machine translation sentences to obtain a system-level score. 
 
Table \ref{tab1} shows a KG-BERTScore calculation example for en-zh language pair. In subsequent experiments, if $\alpha$ parameters in the formula are not described, the default value is 0.5, and if there is no entity in the source, $F_{KG}$ score is 1.

\section{Experiments}\label{sec4}

We conduct a comparative experiment on WMT19 QE as a metric without references shared task to test the effectiveness of our reference-free MT evaluation metric. We use xlm-roberta-base as the default pre-trained multilingual model and the ninth layer of the model for word embedding to calculate $F_{BERT}$ scores.\footnote{\url{https://huggingface.co/xlm-roberta-base}} As for the calculation of $F_{KG}$ score, following the method of Zorik et al\cite{gekhman2020kobe}, we use Google Knowledge Graph Search API to annotate named entities and their entity IDs in sentences.\footnote{\url{https://developers.google.com/knowledge-graph}} The annotated data can be downloaded from http://storage.googleapis.com/gresearch/
kobe/data/annotations.zip.

\subsection{Datasets}\label{subsec11}

We collect the source sentences and system translation sentences from WMT19 news translation shared task, which contains 233 translation systems across 18 language pairs.\footnote{\url{https://www.statmt.org/wmt19/translation-task.html}} Each language pair has approximately 1,000-2,000 source sentences. 

\subsection{Baselines}\label{subsec22}

For each language pair, we apply reference-free BERTScore and KG-BERTScore to score translation systems in WMT19 news translation shared task\cite{barrault2019findings} respectively. We then measure the Pearson correlation of these two scores with human judgements. Finally, we compare a series of reference-free MT evaluation metrics: ibm1-morpheme and ibm1-pos4gram\cite{popovic2012morpheme}, LASER\cite{yankovskaya2019quality}, LogProb\cite{yankovskaya2019quality}, YiSi-2 and YiSi-2-srl\cite{lo2019yisi}, and a reference-based MT evaluation metric: BLEU.

\begin{table}[ht]
\begin{center}
\Huge
\caption{System-level pearson correlation with human judgements for language pairs into English from the WMT19 QE as a metric without references shared task. }\label{tab2}%
\scalebox{0.45}{
\begin{tabular}{@{}lllllllll@{}}
\toprule
src-mt & de-en & fi-en & gu-en & kk-en & lt-en & ru-en & zh-en & mean\\
\hline
BLEU & 0.849  & 0.982  & 0.834  & 0.946  & 0.961  & 0.879  & 0.899  & 0.907 \\
\hline
LASER & 0.247  & - & - & - & - & -0.310  & - & -\\
LogProb & -0.474  & - & - & - & - & -0.488  & - & -\\
ibm1-morpheme & 0.345  & 0.740  & - & - & 0.487  & - & - & -\\
ibm1-pos4gram & 0.339  & - & - & - & - & - & - & -\\
UNI & 0.846  & \textbf{0.930}  & - & - & - & 0.805  & - & -\\
UNI+ & 0.850  & 0.924  & - & - & - & 0.808  & - & -\\
YiSi-2 & 0.796  & 0.642  & -0.566  & -0.324  & 0.442  & -0.339 & 0.940  & 0.227 \\
YiSi-2 srl & 0.804  & - & - & - & - & - & \textbf{0.947}  & -\\
BERTScore  & 0.785  & 0.866  & -0.007  & 0.117  & 0.657  & -0.372  & 0.728  & 0.396 \\
\hline
 KG-BERTScore  & \textbf{0.862}  & 0.733  & \textbf{0.764}  & \textbf{0.936}  & \textbf{0.688}  & \textbf{0.918}  & 0.908  & \textbf{0.830} \\

\bottomrule
\end{tabular}}
\end{center}
\end{table}

\begin{table}[ht]
\begin{center}
\Huge
\caption{System-level pearson correlation with human judgements for language pairs from English from the WMT19 QE as a metric without references shared task. }\label{tab3}%
\scalebox{0.40}{
\begin{tabular}{@{}llllllllll@{}}
\toprule
Metric & en-cs & en-de & en-fi & en-gu & en-kk & en-lt & en-ru & en-zh & mean\\
\hline
BLEU & 0.897  & 0.921  & 0.969  & 0.737  & 0.852  & 0.989  & 0.986  & 0.901  & 0.907 \\
\hline
LASER & - & 0.871  & - & - & - & - & -0.823  & - & -\\
LogProb & - & -0.569  & - & - & - & - & -0.661  & - & -\\
ibm1-morpheme & \textbf{0.871}  & 0.870  & 0.084  & - & - & \textbf{0.810}  & - & - & -\\
ibm1-pos4gram & - & 0.393  & - & - & - & - & - & - & -\\
UNI & 0.028  & 0.841  & \textbf{0.907}  & - & - & - & \textbf{0.919}  & - & -\\
UNI+ & - & - & - & - & - & - & 0.918  & - & -\\
USFD & - & -0.224  & - & - & - & - & 0.857  & - & -\\
USFD-TL & - & -0.091  & - & - & - & - & 0.771  & - & -\\
YiSi-2 & 0.324  & 0.924  & 0.696  & 0.314 & 0.339 & 0.055  & -0.766  & -0.097 & 0.224 \\
YiSi-2 srl & - & \textbf{0.936}  & - & - & - & - & - & -0.118  & -\\
BERTScore & 0.035  & 0.893 & 0.765 & \textbf{0.549} & 0.650 & -0.084 & -0.779  & -0.127 & 0.238 \\
\hline
KG-BERTScore & 0.364 & 0.897 & 0.595 & -0.197  & \textbf{0.839}  & -0.081  & 0.638  & \textbf{0.077} & \textbf{0.392} \\
\bottomrule
\end{tabular}}
\end{center}
\end{table}

\begin{table}[ht]
\begin{center}
\scriptsize
\caption{System-level pearson correlation with human judgements for language pairs excluding English from the WMT19 QE as a metric without references shared task. }\label{tab4}%
\scalebox{1.25}{
\begin{tabular}{@{}lllllllll@{}}
\toprule
Metric && de-cs && de-fr && fr-de && mean\\
\hline
BLEU && 0.941  && 0.891  && 0.864  && 0.899 \\
\hline
ibm1-morpheme && 0.355  && -0.509  && -0.625  && -0.260 \\
ibm1-pos4gram && - && 0.085  && \textbf{-0.478}  && -\\
YiSi-2 && 0.606  && \textbf{0.721}  && -0.530  && 0.266 \\
BERTScore && 0.572  && 0.692  && -0.746  && 0.173 \\
\hline
KG-BERTScore && \textbf{0.959}  && 0.556  && -0.713  && \textbf{0.267} \\
\bottomrule
\end{tabular}}
\end{center}
\end{table}

\subsection{Results}\label{subsec33}

The results for language pairs into English are available in Table \ref{tab2}. Reference-free KG-BERTScore outperforms all other reference-free MT evaluation metrics for de-en, gu-en, kk-en, lt-en and ru-en. The average Pearson correlation of reference-free KG-BERTScore on all language pairs into English is 0.830, only 0.077 lower than that of BLEU. Table \ref{tab3} describes the 
results for language pairs translated from English, reference-free KG-BERTScore outperforms for en-kk and en-zh. Besides, the results for language pairs not involving English are available in Table \ref{tab4}. In this case, reference-free KG-BERTScore outperforms for de-cs with Pearson correlation of 0.959. In conclusion, reference-free KG-BERTScore has a higher overall pearson correlation with human judgements than reference-free BERTScore metric and the other metrics we know for reference-free MT evaluation.

In addition, we also notice that KG-BERTScore does not perform very well on language pairs such as en-gu, en-lt and fr-de, which is due to the insufficient embedding conversion ability of pre-trained multilingual model and the weak named entities coverage of multilingual knowledge graph.

\section{Analysis}\label{sec5}

Summarizing the above findings, reference-free KG-BERTScore obtains the best results on 8 out of 18 language pairs. This means that incorporating multilingual knowledge graph into BERTScore is a promising path towards reference-free MT evaluation. In this section, we analyze the factors that affect the effectiveness of the reference-free KG-BERTScore metric.

\subsection{Impact of Different Pre-training Multilingual Models}\label{subsec111}

\begin{figure}[ht]
\begin{tikzpicture}
    \begin{axis}
    [
        height=6cm,
        width=7cm,
        ybar,
        enlargelimits=0.3,
        legend style={at={(0.5,-0.2)}, 
        anchor=north,legend columns=0.9}, 
        ylabel={\#Average pearson correlation with human judgments for language pairs into English},
        xlabel={\#reference-free BERTScore and KG-BERTScore metrics based on different pre-training multilingual models},
        symbolic x coords={bert-base-multilingual-cased,xlm-roberta-base,xlm-roberta-large},
        xtick=data,
         nodes near coords, 
        nodes near coords align={vertical},
        ]
        \tiny
        \addplot[fill=cyan]  coordinates {(bert-base-multilingual-cased,0.246) (xlm-roberta-base,0.396)  (xlm-roberta-large,0.631)};
        \addplot[fill=orange]  coordinates { (bert-base-multilingual-cased,0.789)  (xlm-roberta-base,0.830) (xlm-roberta-large,0.851)};
        \legend{BERTScore, KG-BERTScore} 
    \end{axis}
\end{tikzpicture}
\caption{Average system-level pearson correlation with human judgements of reference-free BERTScore and KG-BERTScore metrics based on different pre-training multilingual models for language pairs into English.}
\label{fig1}
\end{figure}
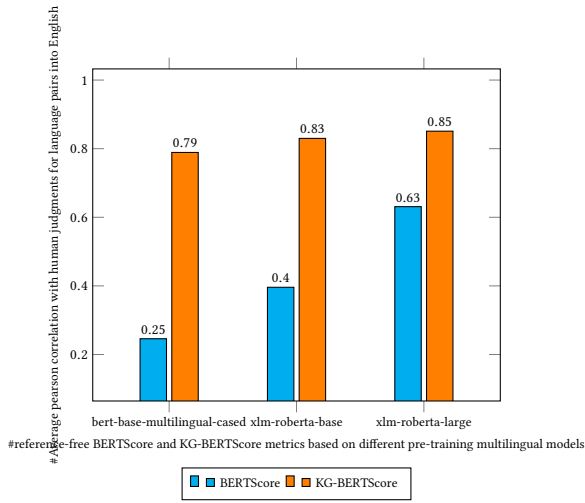

To measure the impact of different pre-trained multilingual models on reference-free BERTScore and KG-BERTScore, we select several commonly used pre-trained l models: bert-base-multilingual-cased,\footnote{\url{https://huggingface.co/bert-base-multilingual-cased}} xlm-roberta-base, xlm-roberta-large.\footnote{\url{https://huggingface.co/xlm-roberta-large}} Based on these l models, reference-free BERTScore and KG-BERTScore are employed to evaluate the language pairs into English from WMT19 QE as a metric without references shared task. The average pearson correlation between the evaluation results of all language pairs and human judgments is shown as Figure \ref{fig1}. 

Experimental results show that while the pre-trained multilingual model performs better on reference-free BERTScore metric, it also performs better on reference-free KG-BERTScore metric. Furthermore, reference-free KG-BERTScore metric is consistently more accurate than reference-free BERTScore metric under the same pre-trained multilingual model.

\subsection{Impact of Different weights in reference-free KG-BERTScore}\label{subsec222}

To explore the impact of different weights in reference-free KG-BERTScore metric, we apply the reference-free KG-BERTScore metrics with weights of 0.0, 0.2, 0.4, 0.5, 0.6, 0.8, 1.0 to evaluate the language pairs into English from WMT19 QE as a metric without references shared task. Table \ref{tab5} indicates the pearson correlation between the evaluation results and human judgments for each language pair. The experimental results show that the combination of knowledge graph and BERTScore is better than that of only knowledge graph or BERTScore, and when the weight is 0.5, the overall evaluation accuracy is close to the best.

\begin{table}[ht]
\begin{center}
\Huge
\caption{System-level pearson correlation with human judgements of reference-free KG-BERTScore metrics with different weights for language pairs into English.}\label{tab5}
\scalebox{0.50}{
\begin{tabular}{@{}lllllllll@{}}
\toprule
src-mt & de-en & fi-en & gu-en & kk-en & lt-en & ru-en & zh-en & mean\\
\hline
$\alpha$=0.0 &  0.785  & \textbf{0.866}  & -0.007  & 0.117  & 0.657  & -0.372  & 0.728  & 0.396 \\
$\alpha$=0.2 & 0.852  & 0.857  & 0.547  & 0.867  & 0.686  & 0.766  & 0.891  & 0.781 \\
$\alpha$=0.4 & 0.861  & 0.774  & 0.739  & 0.926  & \textbf{0.688}  & 0.904  & 0.906  & 0.828 \\
$\alpha$=0.5 & 0.862  & 0.733  & 0.764  & 0.936  & \textbf{0.688}  & 0.918  & 0.908  & \textbf{0.830} \\
$\alpha$=0.6 & 0.864  & 0.696  & 0.774  & 0.943  & \textbf{0.688}  & 0.924  & 0.910  & 0.828 \\
$\alpha$=0.8 & 0.865  & 0.636  & \textbf{0.778}  & 0.950 & \textbf{0.688}  & 0.930  & 0.913  & 0.823 \\
$\alpha$=1.0 & \textbf{0.866}  & 0.592  & 0.776  & \textbf{0.954}  & \textbf{0.688}  & \textbf{0.931}  & \textbf{0.914}  & 0.817 \\
\bottomrule
\end{tabular}}
\end{center}
\end{table}

\section{Conclusion}\label{sec6}

In the paper, a reference-free KG-BERTScore metric is proposed for MT evaluation. Compared with traditional metrics, the metric is unsupervised and does not require parallel corpus and human scoring data for pre-training and fine-tuning, but only requires multilingual knowledge graph and pre-trained multilingual model. We also verify the effectiveness of KG-BERTScore on WMT19 QE as a metric without references shared task, and its experimental results show that the metric is reliable and promising.



\end{document}